\pdfoutput=1
\documentclass[11pt]{article}

\usepackage[]{coling}

\usepackage{times}
\usepackage{latexsym}
\usepackage{comment}

\usepackage[T1]{fontenc}

\usepackage[utf8]{inputenc}

\usepackage{microtype}

\usepackage{graphicx}
\usepackage{booktabs}
\usepackage{dm_utility}
\usepackage{annotation}
\usepackage{cleveref}
\usepackage{hyperref}
\usepackage{enumitem}
\usepackage[title]{appendix}
\usepackage{soul}
\usepackage{graphicx}
\usepackage{enumitem}
\usepackage{makecell}

\newcommand{\apt}[1]{\textit{#1}}
\newcommand{\gpt}{ChatGPT-3.5-turbo-0613}
\newcommand{\gptshort}{ChatGPT}
\newcommand{\pp}[2]{\ul{#1}\textsubscript{(#2)}}
\newcommand{\rapty}{APTY\textsubscript{Ranked}} %
\newcommand{\bapty}{APTY\textsubscript{Base}} %

\tcbuselibrary{listingsutf8}

\global

\tcbset{
  aibox/.style={
    width=474.18663pt,
    top=10pt,
    colback=white,
    colframe=black,
    colbacktitle=black,
    enhanced,
    center,
    attach boxed title to top left={yshift=-0.1in,xshift=0.15in},
    boxed title style={boxrule=0pt,colframe=white,},
  }
}
\newtcolorbox{AIbox}[2][]{aibox,title=#2,#1}

\title{Towards Human Understanding of Paraphrase Types in Large Language Models}%

\author{
  Dominik Meier\textsuperscript{1,2,3}, Jan Philip Wahle\textsuperscript{1}, Terry Ruas\textsuperscript{1}, Bela Gipp\textsuperscript{1} \\
  \textsuperscript{1}University of Göttingen, Germany \\
  \textsuperscript{2}LKA NRW, Germany\\
  \textsuperscript{3}\texttt{meier@gipplab.org}\\
}

\begin{document}

\maketitle
\AddAnnotationRef
\begin{abstract}
Paraphrases represent a human's intuitive ability to understand expressions presented in various different ways.
Current paraphrase evaluations of language models primarily use binary approaches, offering limited interpretability of specific text changes.
Atomic paraphrase types (APT) decompose paraphrases into different linguistic changes and offer a granular view of the flexibility in linguistic expression (e.g., a shift in syntax or vocabulary used).
In this study, we assess the human preferences towards ChatGPT in generating English paraphrases with ten APTs and five prompting techniques.
We introduce APTY (Atomic Paraphrase TYpes), a dataset of 800 sentence-level and word-level annotations by 15 annotators.
The dataset also provides a human preference ranking of paraphrases with different types that can be used to fine-tune models with RLHF and DPO methods. 
Our results reveal that ChatGPT and a DPO-trained LLama 7B model can generate simple APTs, such as additions and deletions, but struggle with complex structures (e.g., subordination changes).
This study contributes to understanding which aspects of paraphrasing language models have already succeeded at understanding and what remains elusive.
In addition, we show how our curated datasets can be used to develop language models with specific linguistic capabilities.
\end{abstract}

\section{Introduction}
   Paraphrases are changes in a text's wording or structure, resulting in a new text with approximately the same meaning \cite{bhagat-hovy-2013-squibs,vilaThisParaphraseWhat2014,wahle-etal-2023-paraphrase}. 
   Paraphrase plays a fundamental role in NLP as understanding the variability in linguistic expression is key for various tasks, e.g.,
   prompt engineering, text summarization, and plagiarism detection \cite{zhou2022large,el2021automatic,barron-cedeno-etal-2013-plagiarism}. 
   Many have assessed whether two texts convey the same meaning through a single similarity score or binary assessment, limiting the granularity of predictions.
   
   Atomic Paraphrase Types (APT) \cite{barron-cedeno-etal-2013-plagiarism,vilaThisParaphraseWhat2014} can be used as a new lens through which the linguistic relationship between two paraphrases can be explained. 
   Generating and detecting APTs over binary categorizing paraphrases has multiple advantages \cite{wahle-etal-2023-paraphrase}. 
   For example, APTs can pinpoint whether a sentence's grammatical structure or the used vocabulary has changed between potential plagiarism cases \cite{Alvi2021}.
   Understanding how language models understand this variation in linguistic expression gives us insights into how their understanding of language differs from that of humans.
   It also explains in which language aspects models are proficient, where challenges remain, and how we can make models more robust to a wide array of paraphrase characteristics (e.g., syntactical and lexical changes).     
   There are many ways in which two paraphrases can differ. Consider the following example:
   
    \begin{minipage}[c]{0.9\linewidth}
            \vspace{2mm}
            \textbf{Original}: ``\pp{They}{a} had published an advertisement on the Internet \pp{on June 10}{b}, offering the \pp{cargo}{c} for sale, \pp{he added}{d}.''\\
            \textbf{Paraphrase}: ``\pp{On June 10}{b}, \pp{the ship's owners}{a} had published an advertisement on the Internet, offering the \pp{explosives}{c} for sale.''
            \vspace{2mm}
    \end{minipage}

    Here, the paraphrase contains the following APT changes: (a) and (c) change the lexical unit for another one with the same meaning, (b) re-orders the words in the sentence, and (d) adds lexical and functional units.
    \begin{figure*}[ht]
    \centering
        \includegraphics[scale=0.75]{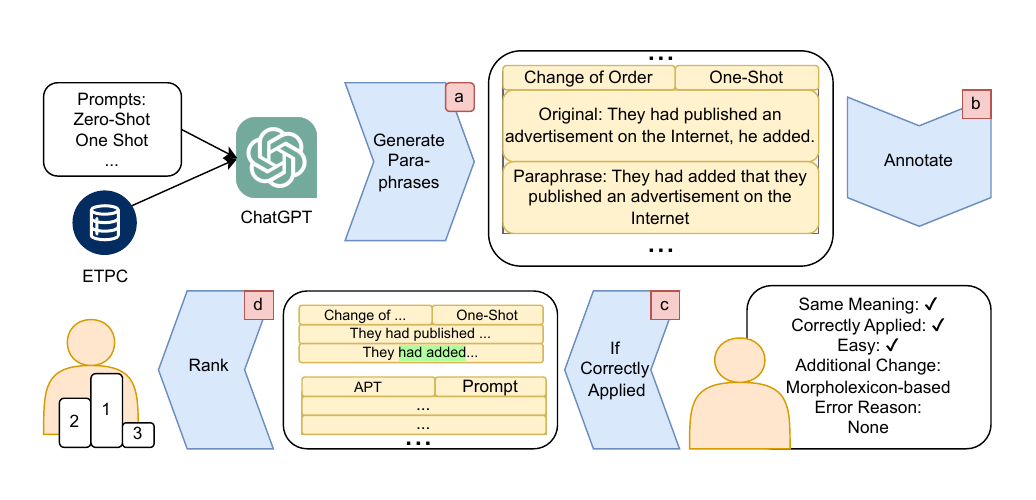}
        \caption{The generation and annotation process. During paraphrase generation (a), we select samples from the ETPC dataset \cite{kovatchev-etal-2018-etpc} and prompt \gpt\space \cite{openai_gpt3.5_turbo_06} using zero-shot, one-shot, few-shot, chain-of-thought, and a fine-tuned model to generate new examples considering selected paraphrase types. For each technique and paraphrase type combination, we sample ten sentences. With five prompting techniques and ten selected APTs, we produce 500 sentence pairs. In (b), paraphrased candidates are annotated by 15 humans, who answer questions and highlight the word spans of the change. In (c), we select the generations in which the APT has been applied correctly, and in (d), the selected generations are ranked from worst to best. }
        \label{fig:annotation_process}
    \end{figure*}
    
    So far, how well models generate or detect paraphrases with specific APTs has been largely unknown.
    In this work, we asked 15 humans to annotate 800 APT generations with various properties such as the perceived difficulty of generation, the model's success at generating a certain type and the reasons behind their failure, its confusion with other types, and the similarity of the APT on a sentence and word level. 
    We publish this dataset as \textbf{A}tomic \textbf{P}araphrase \textbf{TY}pes Base (APTY\textsubscript{Base})\footnote{\href{https://github.com/worta/apty}{https://github.com/worta/apty}}.
    To further contribute to the future of research on APTs, we extended this new dataset by ranking the different APT examples by human preferences that can be used to optimize paraphrase generation models in the future using human preference methods such as RLHF \cite{ouyang2022training} and DPO \cite{rafailov2024direct} called \rapty.
    We demonstrate the usefulness of the preference data by using DPO to train a Llama 2 7B \cite{touvron2023llama} based model and compare the generation performance to a supervised fine-tuned and base version.
    The whole generation and annotation process is shown in \Cref{fig:annotation_process} and discussed in \Cref{sec:methodology}.

    \looseness=-1Our results show that \gpt\space(\gptshort)\space is capable of generating \apt{Same Polarity Substitution}, \apt{Semantic-Based Changes}, and \apt{Change of Order} and struggle with generating \apt{Inflectional Changes}, \apt{Synthetic/Analytic Substitutions}, and \apt{Subordination and Nesting Changes}. 
    In general, changes requiring deeper grammatical understanding are difficult to generate. %
    Few-shot and chain-of-thought (CoT) prompting have increased generation success compared to other prompting techniques, especially for \apt{Addition/Deletion} and \apt{Semantic-Based Changes}.
    Surprisingly, humans often rank CoT generations lower than other methods.
    We also found that the most common error for \gptshort\space when applying APT is applying the wrong kind of APT and that morpholexical changes, i.e., changes that arise at the word or morpheme level, are the most commonly wrongly or additionally applied.
    Lastly, the DPO-trained LLama outperforms the supervised fine-tuned and base Llama models by a wide margin. 
    Our main contributions are:
    
    \begin{enumerate}[topsep=0.5pt, itemsep=0pt]
        \item A human study with 500 annotations of 15 participants on the ability of \gptshort\space to \hyperref[q_apt_generation]{generate paraphrase types (Q1)};
        \item A new dataset (\bapty) with sentence-pair information on sense-preservation, specific paraphrase type applied, the location of the change, and error reasons over \hyperref[q_apt_generation]{five methods of paraphrase generation (Q2)}, with different prompt styles of zero-shot, one-shot, few-shot, chain-of-thought, and a fine-tuned model;
        \item Analysis of \hyperref[q_quality]{human preferences (Q3)} of paraphrase generation and a new dataset \rapty \space with human-ranked paraphrase type generations using best-worst scale%
        \item Investigation of \hyperref[q_mistakes]{types of errors (Q4)} \gptshort\space makes when generating APTs and \hyperref[q_confusion]{how much \gptshort\space confuses APTs (Q5)}, i.e., confusion between the types;
        \item Evaluation of \hyperref[q_dpo]{DPO training (Q6)} with the obtained human preferences based on the LLama 2 7B model family with 300 annotations
    \end{enumerate}

\section{Related Work} 
    Approaches for paraphrase generation range from rule- and template-based approaches  \cite{androut2010survey} to trained transformers generating paraphrased text \cite{wahle-etal-2022-large}.
    Rule-based methods rely on parsing the original sentence and applying either hand-crafted \cite{keownParaphrasingQuestionsRule} or automatically inferred \cite{lin2001discovery} rules to transform the text. 
    Recently, paraphrase generation involves deep learning models, especially LLMs \cite{zhou-bhat-2021-paraphrase}.
    \citet{witteveen-andrews-2019-paraphrasing} use a fine-tuned version of GPT-2 to generate paraphrases and evaluate their semantic similarity. 
    \citet{PALIVELA2021100025} fine-tune a T5 model to generate and identify paraphrases. 
    \citet{wahle-etal-2022-large} explore T5 and GPT-3 regarding qualitative properties of generated paraphrases,  access the ability of humans to identify machine-paraphrased text, and suggest that LLMs can generate paraphrases that match human-generated paraphrases in clarity, fluency, and coherence.

    As previous paraphrase tasks rely heavily on similarity scores and do not capture the linguistic flexibility of paraphrases, \citet{wahle-etal-2023-paraphrase} proposed two new tasks, i.e., paraphrase type generation and detection, using the ETPC \cite{kovatchev-etal-2018-etpc} dataset.
    Their findings indicate that current LLMs (e.g., \gptshort) perform well when generating paraphrases with generic semantic similarity but struggle to generate them with fixed APTs.
    Additionally, models trained with APTs have improved performance in general paraphrase tasks (i.e., without APTs). While little is known yet in full detail on the paraphrastic mechanisms of all paraphrase types, further work revealed that specific types elicit prompt engineering capabilities over various downstream tasks, e.g., polarity for sentiment, or discourse for summarization \cite{wahle2024elicitprompt}.
    An alternative approach to APT to have more control over linguistic properties is generating paraphrases based on syntactic templates \cite{huang2021generating}. However, changes on a template level are too specific in a generation or detection task for humans to specify desired concepts. 
    
    Although \citet{wahle-etal-2023-paraphrase} has sparked interest in more granular paraphrasing, their work relies only on automatic metrics such as BLEU \cite{10.3115/1073083.1073135} or ROUGE \cite{lin-2004-rouge}, which is limited and lacks the human component in the evaluation process.
    We explore human preferences in paraphrases by examining whether a paraphrase type was correctly applied by a model, the kind of errors made by \gptshort\space at the time of generation, and by ranking paraphrases according to human judgement.

\section{Methodology}
    \looseness=1Our experiments are split into two parts, i.e., generating paraphrase types using \gptshort\space and annotating and evaluating their outputs according to multiple criteria with human participants.
    The process is shown in \Cref{fig:annotation_process}.
    In the following subsections, we detail the generation and annotation processes. 

    \subsection{Paraphrase Type Generation} \label{sec:methodology}
    We consider a variant of the paraphrase type generation task described in \cite{wahle-etal-2023-paraphrase}.
    Given an APT $l \in L$, where $L$ is the set of possible paraphrase types, and $x$ a base sentence, we want to generate a paraphrase $\tilde{x}$ incorporating the given change $l$ while maximizing the similarity between $x$ and $\tilde{x}$. 
    We limit ourselves to applying a single paraphrase type for the highest degree of control; we leave research into the complexity of combining different APTs for future work.
    We do not restrict the position where the change has to be applied; the model must choose where to apply the change.

    \looseness=-1In our experiments, we use the ETPC dataset \cite{kovatchev-etal-2018-etpc}, which contains pairs of original and paraphrased sentences annotated with APTs.
    We choose the ten APTs with the most examples in the dataset as primary for this study. 
    We excluded \apt{Identity} changes, as those contain no change in the sentence; \apt{Same Polarity Substitution (habitual and named entity)} as it is similar to \apt{Same Polarity Substitution (contextual)}, which could harm the diversity for the chosen paraphrase types.
    We also exclude \apt{Syntax/Discourse Structure} changes as it would require our annotators to understand all other paraphrase types, even those not considered in the study. 
    \Cref{app:full_list} details a full list of types.

    We sample ten paraphrase pairs for each paraphrase type $l$. 
    We use the first sentence in a paraphrase pair as a base sentence and generate the paraphrase using \gpt. 
    The prompts are constructed by asking the model to generate a sentence with the same meaning using the APT $l$. 
    We prompt the model using zero-shot, one-shot, few-shot \cite{NEURIPS2021_8493eeac}, chain-of-thought (CoT) \cite{weiChainofThoughtPromptingElicits2023}, and a fine-tuned model from \citet{wahle-etal-2023-paraphrase}, which is based on \gpt\space as well.
    The prompt contains the name of type $l$ and its definition, taken from \cite{barron-cedeno-etal-2013-plagiarism} with minimal changes to fit the prompt (see \Cref{sec:prompts} for exact prompts used.)
    Similarly to \citet {NEURIPS2021_8493eeac}, five few-shot examples are given; one instance with added reasoning is provided for CoT prompts. \Cref{sec:apt_def} details the exact APT definitions.
    We use the same base sentences for evaluating the DPO-tuned LLama model and follow the prompt template for the fine-tuned model. We use Llama 2 7B, a version of LLama 2 7B fine-tuned on the ETPC dataset from \citet{wahle-etal-2023-paraphrase}, and a DPO-trained Llama model trained by us on the ranked data from the second phase of the annotation and based on the fine-tuned model.

    \subsection{Annotation}
    \label{sec:annotation}
     We recruited 15 annotators to perform ratings in two phases. 
     The annotators were students from computer science, data science, and related programs with self-accessed English language skills of C1 or C2 and an average age of 24. 
     In two one-hour meetings, they were trained with detailed guidelines. \Cref{app:add_anno_info,app:guidelines} detail the annotation process and the guidelines.
     
     In the first phase, the annotators were shown a pair of original and paraphrased candidates and APTs to be applied. 
     The annotators were asked whether the given sentence pair had approximately the same meaning,  to specify if a specific APT was applied correctly, and to determine the groups of any additional APT if multiple changes were made to the original sentence or if the given APT was not applied correctly.
     Annotators were also asked to highlight the word position of change if the paraphrase type had been applied correctly (i.e., a span of multiple words that can be consecutive or disjoint).
     \Cref{fig:anno_example} in \Cref{app:guidelines} shows and example of the system.
     We also included five manually created gold examples to certify the annotations were carried out carefully and to check for agreement among annotators; no annotator answered more than one gold question incorrectly; the details can be found in \Cref{sec:appex}.
     On average, the median time to complete the annotation of one paraphrase pair was 74 seconds, with the 25th percentile being 47 seconds and the 75th percentile 121 seconds.
     Each paraphrase type was annotated by one annotator, except for our gold examples, which were given to all annotators. 
     These annotations compose the \bapty \space dataset.

     In the second phase, paraphrases with successfully applied paraphrase types were ranked from best to worst \cite{flynn2014best}.
     The list was discarded if less than two generations were successful for a given sentence. 
     In total, 80 of 100 possible lists remained, and the annotators were then asked to rank them according to their preferences.
     Five annotators gave their preferences for each list, leading to $5 \times 80 = 400$ preference annotations composing the \rapty \space dataset.

     For the evaluation of the DPO-tuned Llama model, two annotators performed the first phase again, but each annotated every one of the 300 (3 models x 10 APT x 10) generations.

\section{Experiments} \label{sec:experiments}
        \question{How successful is \gptshort, on average, at generating specific paraphrase types? Which paraphrase types can it already produce with high success, and which ones do they struggle with?}
        \label{q_apt_generation}
        \short %
        \apt{Change of Order}, \apt{Semantic Changes}, and \apt{Same Polarity Substitution} can be generated with decent success rates, while more complicated changes, particularly those involving deeper grammatical changes at multiple points like \apt{Derivational Changes} pose a problem.

        \ans
        Based on the annotation of the first phase, we measure the success rate of the different prompt and paraphrase type combinations to assess which APTs \gptshort\space could already generate well and which APTs are more challenging.
        The results are shown in \Cref{fig:success}. %
        We color the different generation strategies and group them by APT on the x-axis.
                
          \begin{figure}[t]
            \centering
            \includegraphics[width=\linewidth]{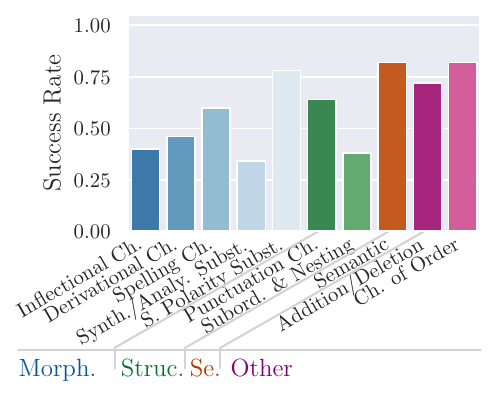}
            \caption{The success rate on the y-axis in generating the specific APT on the x-axis across all tested prompting techniques of \gptshort.}
            \label{fig:averages}
        \end{figure}  
        Humans often judge the generation as successful when prompting \gptshort\space to generate \apt{Change of Order} (82\%), \apt{Semantic Changes} (82\%), and \apt{Same Polarity Substitution} (78\%).
        In contrast, \apt{Derivational Changes} (46\%), \apt{Subordination and Nesting Changes} (38\%), and \apt {Synthetic/Analytic Subsitution} (34\%) show low success rates.
        While \apt{Change of Order} and \apt{Same Polarity Substitution} require small changes and understanding on the sentence level or of grammatical concepts, \apt{Derivational Changes} require an understanding of the parts-of-speech of the affected lexical unit.
        \apt{Subordination and Nesting Changes} similarly require a nuanced understanding of sentence structure and grammar.  
        Consider the following example of a paraphrase using both a \apt{Derivational change} (a) and  \apt{Same Polarity Substitution} (b).

    \begin{figure*}[!t]
    \centering
    \includegraphics[width=\linewidth]{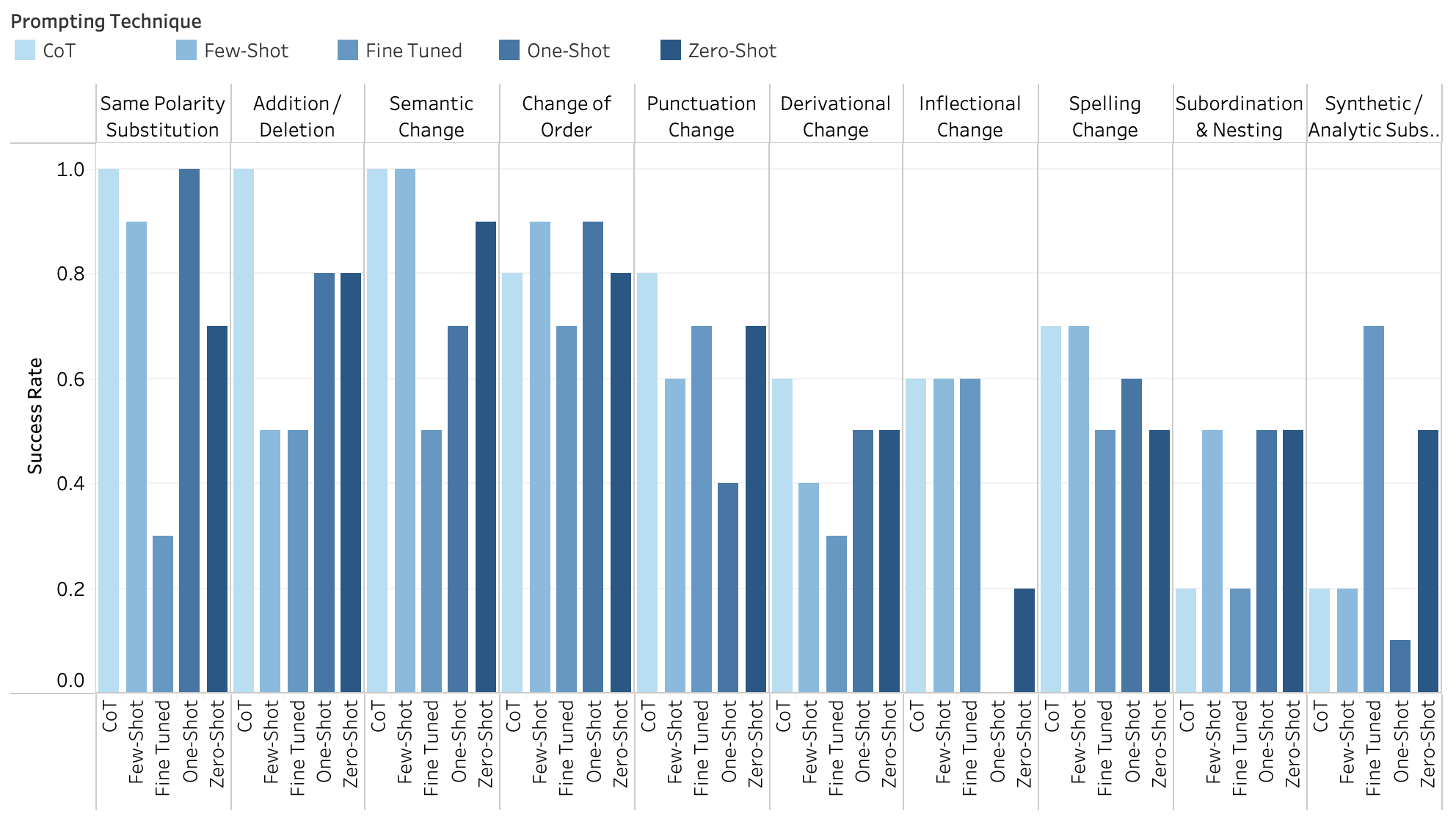}
    \caption{The success rate in generating a specific APT for different prompting techniques of \gptshort.}
    \label{fig:success}
\end{figure*}
        
       \begin{minipage}[c]{0.9\linewidth}
            \vspace{4mm}
            Original: ``A \pp{smiling}{a} senior \pp{policeman}{b} shook hands with Mr Laczynski''\\
            Paraphr.: ``A senior \pp{police officer}{b} \pp{smiled}{a} and shook hands  with Mr Laczynski'' 
            \vspace{4mm}
        \end{minipage}

        Note that for \apt{Same Polarity Substitution}, no change in sentence structure or on word level somewhere else in the sentence is necessary, i.e., the change is locally constrained and does not require understanding beyond synonym relations. 
        However, changing \textit{smiling} to \textit{smiled} makes it necessary to shift the word's position, as it is now used as a verb. 
        It needs to be conjugated to match the rest of the sentence, which is in the simple past tense and third person singular and requires the addition of \textit{and} to keep the sentence correct.
        The high success rate of 82\% for \apt{Semantic Changes} is surprising as a semantic shift is typically complex.
        But the change also often involves idioms and other common turns of phrases, which might be seen often in training data.
        Structure-based changes present further difficulty for the model. %
        These changes require a deep grammatical understanding of the text.

        In the general paraphrase generation task without APTs, models might seem to generate diverse paraphrases but repeatedly use a small selection of APTs that convince annotators of diversity as a form of reward hacking \cite{casper2023open}.  
        One explanation could be that LLMs are more familiar with some paraphrase types than others, i.e., paraphrasing is also biased towards certain types in the training data and fine-tuning steps \cite{zhou2022paraphrase}.
        Without diversity of paraphrase types in training, models might be restricted in linguistic diversity, even in cases where underrepresented paraphrase types would be advantageous, similar to how it hinders them in detection tasks \cite{zhou2022paraphrase}.

        \question{How do different prompting techniques affect the success in generating paraphrase types?} %
        \label{q_prompt_styles}
        \short CoT is the most successful prompting strategy, while one-shot prompting led to the highest error rate.

        \ans \Cref{fig:success} decomposes \Cref{fig:averages} into individual prompting techniques to compare how different prompt techniques influence the success rate of \gptshort\space in generating paraphrase types. 
        This allows us to observe how familiar \gptshort\space is with the tasks (zero-shot performance) and how much reasoning of \gptshort\space improves performance (CoT). 
        For detailed values, refer to \Cref{sec:appex} \Cref{tab:detail_groups}.
        \looseness=1 Depending on the APT, prompt and type combinations have notable differences in success rates.
        For example, while CoT is relatively successful at \apt{Same Polarity Substitution} when zero-shot prompted, \gptshort\space struggles with this task. 
        
        Human annotators found that the CoT prompt was applied successfully in 69\% of the cases, followed by few-shot (63\%), zero-shot (61\%), and one-shot prompt (55\%).
        The fine-tuned model had the worst performance with only 50\%.
        As the fine-tuned model is trained on applying multiple APTs at once, applying a single change might have compromised its performance.
        The one-shot prompt might overly rely on the given example and perform worse on APTs where different changes belong to the same APT. 
        For instance, a verb might be changed to an adjective in the one-shot prompt for generating \apt{Derivational Changes}.

        Our findings suggest that providing multiple examples of the same type and reasoning about the generation improves APT generation performance in LLMs. 
        For some examples, \gptshort\space profits from reasoning about the task at hand, but often, the performance remains poor (e.g., \apt{Synthetic/Analytic Substitution}), suggesting a need to improve model understanding.

        \question{Are there qualitative differences between prompt methods? Do humans favor the generation of certain prompt styles more than others?}
        \label{q_quality}
        \short Annotators favor the generations from few-shot prompting the most, while generations from CoT and the Fine-Tuned model are generally preferred less.

        \ans Besides whether a change is applied correctly, the resulting paraphrases have varying degrees of quality.
        Humans might prefer one paraphrase over another, although they have applied the same type (e.g., some grammatical structures are easier to grasp than others). 
        These preferences can indirectly inform what criteria humans use to judge paraphrases, e.g., at which position a change happens or how creative a change is.
        We investigate whether different prompt techniques (e.g., CoT, zero-shot) produce paraphrases rated as more or less favorable by humans. 
        
        Of all paraphrases where humans have annotated that the APT was correctly applied, we ask annotators to rank these paraphrases from best to worst according to their preferences. %
        The best paraphrase is ranked first, i.e., lower numbers are better.
        To avoid penalizing prompt methods that produced fewer successful generations, we did not rank unsuccessful generations, and we only looked at the rate at which a generation from one prompt technique was preferred compared to another when both generated an APT successfully.
        We give the average ranks of the different prompt techniques and the proportion of times one generation is preferred to another in \Cref{tab:favored_rate}.
        Based on the average rank, few-shot and one-shot generated the most preferred paraphrases, zero-shot and CoT seem to be in the middle, and fine-tuned is ranked worst. 
        For the direct comparison, the trend repeats; we see that few-shot is generally favored compared to the other approaches and fine-tuned and CoT are often perceived as worse.

        As generations from few-shot prompts are perceived best on average, it suggests that practical examples allow \gptshort\space to generate natural-sounding paraphrases more than reasoning.
        Further, while CoT has one of the highest success rates in generating APTs, it seems to generate paraphrases of lesser quality.
        One reason for the poor reception of CoT-generated paraphrases might be that more consideration of the formal aspects of a paraphrase leads to less natural outcomes, which are not perceived well by the annotators.
        The generations of the fine-tuned model were also rated lower than prompting the general model.
        We noted that the fine-tuned model struggled to generate specific APT changes; this seems to extend to natural-sounding paraphrases with single APTs.
        That highlights the importance of high-quality datasets tailored to the specific tasks for fine-tuned models.
        Our preference data can be used to improve model generations via techniques like RLHF or DPO. %
       \setlength{\tabcolsep}{2pt}
        \begin{table}[t]
            \centering            \begin{tabular}{lcccccc}
    \toprule
    Prompt & CoT & \makecell{Few- \\ Shot} & \makecell{Fine \\ Tuned} & \makecell{One-\\ Shot} & \makecell{Zero- \\ Shot} &\makecell{Avg. \\ Rank} \\    
    \midrule
    CoT & 0.00 & 0.27 & 0.63 & 0.33 & 0.40 & 2.27 \\
    Few-Shot & 0.55 & 0.00 & 0.66 & \textbf{0.44} & 0.55 & \textbf{1.84} \\
    Fine-Tuned & 0.34 & 0.29 & 0.00 & 0.28 & 0.37 & 2.54 \\
    One-Shot & \textbf{0.58} & \textbf{0.38} & \textbf{0.68} & 0.00 & \textbf{0.57} & 1.89 \\
    Zero-Shot & 0.54 & 0.32 & 0.59 & 0.33 & 0.00 & 2.21 \\
    \bottomrule
\end{tabular}

            \caption{Proportion of times the generation of the row technique is preferred to the column technique.}
            \label{tab:favored_rate}
        \end{table}

        \question{Which mistakes does \gptshort\space make when generating specific paraphrase types? Does it re-generate identical sentences, apply other types than those instructed, or change the meaning?}
        \label{q_mistakes}
        \short   The most common type of wrong application concerns the model making morpho-lexical changes instead of the requested changes and the boundaries between different types of the same group seem larger than between different groups.  Applying the wrong APT is the most frequent reason for error by far. 

        \ans To determine what kind of errors the models are most likely to make, we asked the annotators to indicate why a paraphrase type was wrongly applied and if other changes were applied instead. 
        If an annotator judged the generated APT from \gptshort\space as a mistake, we provided four possible explanations: the sentences were identical, a different change was applied, the generation was nonsensical, and other reasons.
        The relative frequencies of these error types are listed in \Cref{tab:error_rates}. 
        Applying the wrong type is the most frequent reason for the error, which happens in 60\% of the erroneous cases, while not performing any change is also a common source of error in roughly 20\% of the wrong annotations. 
        Only 17\% of wrongly applied APTs change the meaning from the original sentence, meaning a paraphrase is usually generated even if \gptshort\space fails to use the given APT. 
        An open question is whether giving the model a way to refuse to produce a paraphrase in the prompt if it is unsure, i.e., if it could express uncertainty, would reduce the number of erroneous generations.
        We leave the investigation of this sub-question to future work.
        
        \gptshort\space is good at generating paraphrases with the same meaning but fails to understand the underlying linguistic properties involved.
        It rarely changes the meaning but often changes the wrong aspect of the sentence or does not change the sentence at all.
        Since applying a different type than instructed is the most common issue, we investigate which types are most often mistaken or applied in the following question.

        \begin{table}[t]
            \vspace*{3mm}
            \centering
            \begin{tabular}{lr}
            \toprule
                 Error & Rate in \%  \\
                 \midrule
                 Identical sentences & 21.3 \\
                 Wrong type applied & 60.0 \\
                 Nonsense & 9.9\\
                 Other & 9.9\\
                \bottomrule
            \end{tabular}
            \caption{Error types and their occurrence in percent given incorrect application of an APT.}
            \label{tab:error_rates}
        \end{table}

        \question{Which paraphrase types is \gptshort\space confusing most with another? Do we see a correlation between certain paraphrase types?} %
        \label{q_confusion}
        \short We see intra-group errors less often than inter-group errors and morpho-lexical changes are applied erroneously most often.
        
        \ans 
        The previous answer shows that \gptshort\space often applies different APTs if an error occurs.
        \gptshort\space also applies changes besides the desired one in the case of successful applications.
        In such instances, we can examine the relationship between the different APTs from the perspective of \gptshort, assessing whether and how they correlate. 
        For these cases, the annotators indicated which kind of APT groups these erroneous or additional changes belong to. 
        We used four major groups for annotators to indicate confusion: morpho-lexical, semantic, structural, and other changes; the definitions can be found in the \Cref{sec:apt_groups}.
        
\begin{figure}
        \centering
        \includegraphics[width=\linewidth]{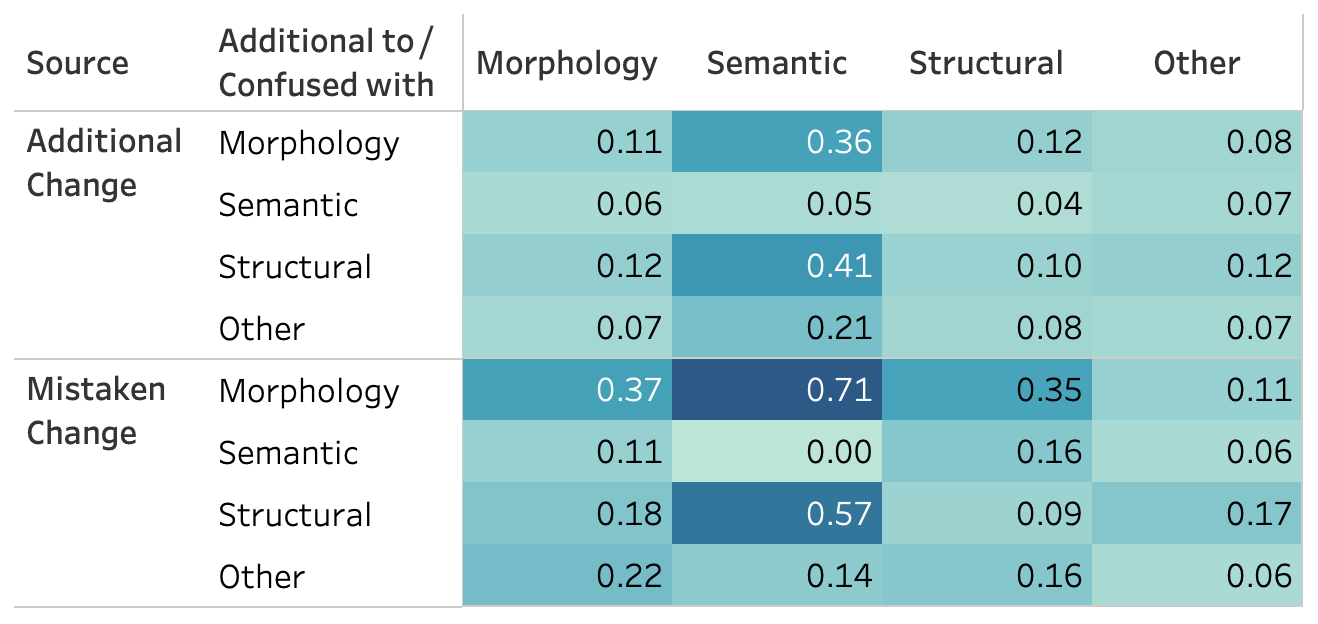}
        \caption{Confusion matrices for additional and erroneous changes. The column gives the intended APT, the row the additional or wrongly applied APT.} %
        \label{fig:confusion}
        \end{figure}

        We show the confusion matrices in \Cref{fig:confusion}, where 'Additional Change' indicates cases where the correct APT is applied along with one or more additional APTs. 
        In contrast, `Mistaken Change' illustrates instances where the model erroneously applies an APT.
        For the numbers in \Cref{fig:confusion}, we count the number of additional erroneous APT applications that belong to the group given at the x-axis, where an APT that is part of the group at the y-axis should have been applied.
        Then, we divide this count by the total number of paraphrases annotated in the group.

        For morpho-lexical changes, the additional changes are evenly spread and happen comparatively rarely. 
        If \gptshort\space makes an erroneous change, it is often a different morpho-lexical change. 
        This is also the case for the other groups, e.g., the most common type of wrong application concerns the model making morpho-lexical changes instead of the requested changes.
        These are probably the most straightforward changes to make without changing the meaning of the sentence and also the most common change, e.g., the most common types in the ETPC dataset are morpho-lexical changes. 
        Generally, the boundaries between different types of the same group seem larger than between different groups. 
        One reason might be that models rarely see isolated changes in the training data (i.e., sentences that only differ by a single APT), leading to a poor understanding and associating them with other changes.
        We also found that the confusion correlates with the difficulty of the paraphrasing task; the more difficult the task is, the more likely it is that other changes are made either in addition or erroneously. 
        We provide more detailed investigations in the additional research questions in \Cref{sec:appex}.

        \question{Can the human preference data improve the generation of APTs?}
        \label{q_dpo}
        \short Training a model with the collected preference data improves the performance of the model. We can see similar trends in LLama as we see in \gptshort.

\begin{figure}
            \centering
            \includegraphics[width=1\linewidth]{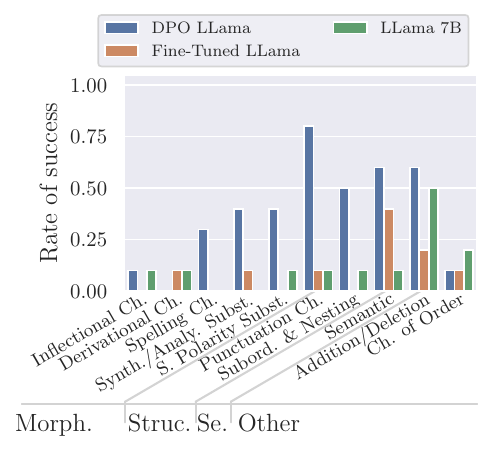}
            \caption{The success rate in generating a specific APT for different LLama 7B Models}
            \label{fig:llama}
        \end{figure}

        \ans We used the collected preferences to train a LLama3 7B model 
        \cite{dubey2024llama}, which was fine-tuned on the ETPC dataset from \cite{wahle-etal-2023-paraphrase}, via DPO \cite{rafailov2024direct}.
        We compare the results in generation success across LLama models (base, fine-tuned, DPO-trained); the results for each APT are shown in \autoref{fig:llama}. 
        Across all generations, DPO performs best with a rate of 34\% of successful APT applications, followed by the base model with 14\% and the fine-tuned model with 11\%.        
        We can also see that, while success rates are generally much lower than with the much larger model, the trends from ChatGPT also hold across the tested LLama 7B variants, i.e., success rates across all variants are best for types where changes are local and where the model has freedom, like \apt{Semantic Changes} (37\%), and \apt{Addition/Deletion} (43\%). 
        At the same time, they are worse for changes that require a deeper grammatical understanding like \apt{Derivational Changes} (8\%).
        We see that the DPO training improves generation success by a large margin, and the collected preference data improves the understanding of the model significantly. 
        The success rates for difficult changes like \apt{Synthentic/Analytic substitution} (10\% to 40\%) or \apt{Subordination and Nesting Changes} (10\% to 50\%), e.g., changes which we have seen ChatGPT struggles to perform as well, received a big boost via DPO training, suggesting an increased understanding of the model of these concepts.
        Surprisingly, fine-tuning on the ETPC dataset does not improve generation success in this task.
        One explanation might be that in ETPC multiple changes are performed at once, and the model does not learn the specific mechanics of a single change due to the notice introduced by other changes.

    \section{Final Considerations}
    \looseness=-1In this work, we contributed to the understanding of how LLMs can generate variability in linguistic expression through paraphrase types.
    Using different prompting techniques, we generated 800 paraphrase candidates. 
    We asked 15 participants to annotate these paraphrase candidates according to whether the paraphrase type was correctly applied by \gptshort\space,  and the kind of errors made by \gptshort\space at generation.
    We made the annotations and human preference rankings available through two new datasets (i.e., \bapty and  \rapty).
   
    We found that CoT prompting generally outperforms other methods for generating paraphrases with specific APTs.
    This suggests that asking \gptshort\space to reason about complex paraphrase types (e.g., \textit{Synthetic/Analytic Substitution}) is ineffective, as the required understanding of these APTs is missing.
    When \gptshort\space made mistakes or added additional changes, they most commonly chose APTs from the morpholexical changes.
    This is likely because these changes are the most common and dominate paraphrasing datasets.
    We also found that for non-morpholexical changes, \gptshort\space rarely confuses changes that can be categorized under the same group (e.g., confusion of one structural change with another instead of a non-structural change).
    One reason might be that \gptshort\space pays increased attention to the properties related to the change so that the retaining of other properties is weighed less.
    \rapty\space opens an interesting avenue for further research in improving the generation and identification quality of APTs using techniques such as DPO or RHLF. 
    Our experiments with Llama 2 7B-based models show a marked advantage when using human preference data from \rapty and DPO and show similar trends to ChatGPT.
    Alternatively, our dataset can be used as a benchmark to develop metrics that better correlate with human preferences.

    \section*{Limitations}
    Although we extend the research in paraphrase types, this study has a few limitations.
    We consider only a selection of APTs and could not investigate all described types due to the resource-intensive work of human annotation.
    We focused on the most common ones in the ETPC dataset, which should give a decent proxy for the most relevant types and selected a sample to ensure diversity in the examined types.
    Additionally, we are limited to one annotation per task for the questionnaire due to effort risking a big variance. 
    However, we used gold examples annotated by each annotator to check the agreement between annotators and to ensure that each annotator performed the annotation attentively.
    We focused on generations from \gpt, as it is one of the most used models, and this is the first investigation of APT paraphrase generation combined with instruction-trained LLMs. 
    We evaluated DPO only for a Llama 7B model; while more models could be interesting, the existing evaluation already shows a marked value of using our preference data.
    Including additional and larger models would make the human evaluation expensive, so we leave this investigation to future work.

\bibliography{cleaned_bib} 
\cleardoublepage
\appendix

\section{Additional Annotation Information}
\label{app:add_anno_info}
 The annotators were paid the usual rate for student workers, corresponding to at least minimum wage in Germany. 
 Consent was explicitly requested via email to publish annotation data and aggregated demographic data.
 No ethics review board was consulted as we judged the collected data as unproblematic.
 The resulting datasets will be released under CC BY 4.0.

Besides the questions mentioned in \ref{sec:annotation}, the annotators also indicated whether additional text was generated, i.e., whether the model generation contained more than just the indicated paraphrase. 
The annotators were instructed to ignore additional text if it could be clearly separated from the paraphrase candidate. For example, sometimes the generated text might be in quotes, or the model might add a phrase like "The paraphrased sentence is:" in front of the generation. 
The question was asked to clean up the data and provide a better foundation for future work.
  
For annotation phase one, we also used gold questions to evaluate whether the annotators understood the assignment and checked their agreement.  
The gold examples include completely correct applied paraphrase types, a correctly applied paraphrase with additional applied paraphrase types, a non-paraphrase, and a paraphrase with a different applied APT. 
The Fleiss Kappa for sense-preservation was 0.92, for correct application 0.68, and for failure reasons 0.61.
Determining if a type was applied correctly can depend on details that can be missed, similarly failure reasons are not always trivial to see and depend on the previous answers.
There was no annotator whose answers to the gold questions suggested a systematic lack of understanding besides normal human errors, e.g., no annotator answered more than one sense-preserving question incorrectly. 
For the second annotation phase, i.e., the preferences, Kendall's coefficient of concordance is 0.52, indicating a moderate agreement on preferences across annotators. 
Precise preferences are likely highly individual, but trends exist.
The Fleiss Kappa for correct application is 0.80 for the DPO annotation with two annotators, showing high agreement.

\subsection{Full List of Considered APT}
\label{app:full_list}
After filtering, the considered paraphrase types are: 
\begin{itemize}[itemsep=0.5pt]
    \item \apt{Addition/Deletion},
    \item \apt{Same Polarity Substitution (cont.)},
    \item \apt{Syntheticanalytic Substitution},
    \item \apt{Change of Order},
    \item \apt{Punctuation Changes},
    \item \apt{Spelling Changes},
    \item \apt{Inflectional Changes},
    \item \apt{Subordination and Nesting Changes},
    \item \apt{Semantic-based Changes},
    \item and \apt{Derivational Changes}.
\end{itemize}

\subsection{Dataset}
The given features in the datasets are given in \Cref{tab:bapty} for \bapty\space and in \Cref{tab:rapty}\space for \rapty. For \rapty\space we give the ranking pairwise and follow the format of Anthropic for RHLF \cite{bai2022training} and add information so that the full information can easily be reconstructed.
\begin{table}[]
    \centering
    \begin{tabular}{lr}
    \toprule
    Attribute & Example \\
    \midrule
    meta & \\
    \quad id & 106 \\
    \quad annotator & 5 \\
    generation & \\
    \quad APT & AdditionDeletion\\
    \quad Kind & One-Shot \\
    \quad original & They had \dots \\
    \quad paraphrase-text & They had \dots \\
    annotation & \\
    \quad paraphrase & True \\
    \quad applied-correctly & True \\
    \quad correct-format & True \\
    \quad hard & False \\
    failure & \\
    \quad identical & False \\
    \quad other & False \\
    \quad nonsense & False \\
    \quad otherchange & False \\
    additional & \\
    \quad morph & False \\
    \quad struct & False \\
    \quad semantic & False \\
    \quad other & False \\
    mistaken & \\
    \quad morph & False \\
    \quad struct & False \\
    \quad semantic & False \\
    \quad other & False \\
    marked-text & \\
    \quad start & 97\\
    \quad end & 109\\
    \quad text & additionally\\
    \bottomrule
\end{tabular}

    \caption{Features of the annotation for \bapty.}
    \label{tab:bapty}
\end{table}

\begin{table}[]
    \centering
    \begin{tabular}{lr}
    \toprule
    Attribute & Example \\
    \midrule
    meta & \\
    \quad id & 100\\
    \quad annotators & [8,7,11,12,14] \\
    original & They had \dots \\ 
    pairwise & \\
        \quad chosen & They had \dots\\
        \qquad ranks & [1,1,2,3,1] \\
        \qquad id & 106 \\
        \quad rejected & Adding that \dots\\
        \qquad id & 104 \\
        \qquad ranks &  [3,4,4,5,4] \\
    \bottomrule
\end{tabular}

    \caption{Features of the annotation for \rapty.}
    \label{tab:rapty}
\end{table}

\section{Additional Experiments and Details}

\subsection{Additional Questions}
  \label{sec:appex}
    \questiona{How does \gptshort\space perform on examples perceived as easy or hard by humans?}%
        \label{q_difficulty}
        \ans 
        The few-shot approach works better in cases where humans evaluate the task as
        hard, while the CoT-prompted model performs poorly for hard examples
    
        Applying different APTs can be challenging for humans, depending on the precise APT and the base sentence. 
        We explore if the human perceived difficulty of paraphrases also relates to the performance of \gptshort; that is, are paraphrasing tasks difficult for humans also difficult for \gptshort?
        
\begin{figure}
            \centering
            \includegraphics[width=\columnwidth]{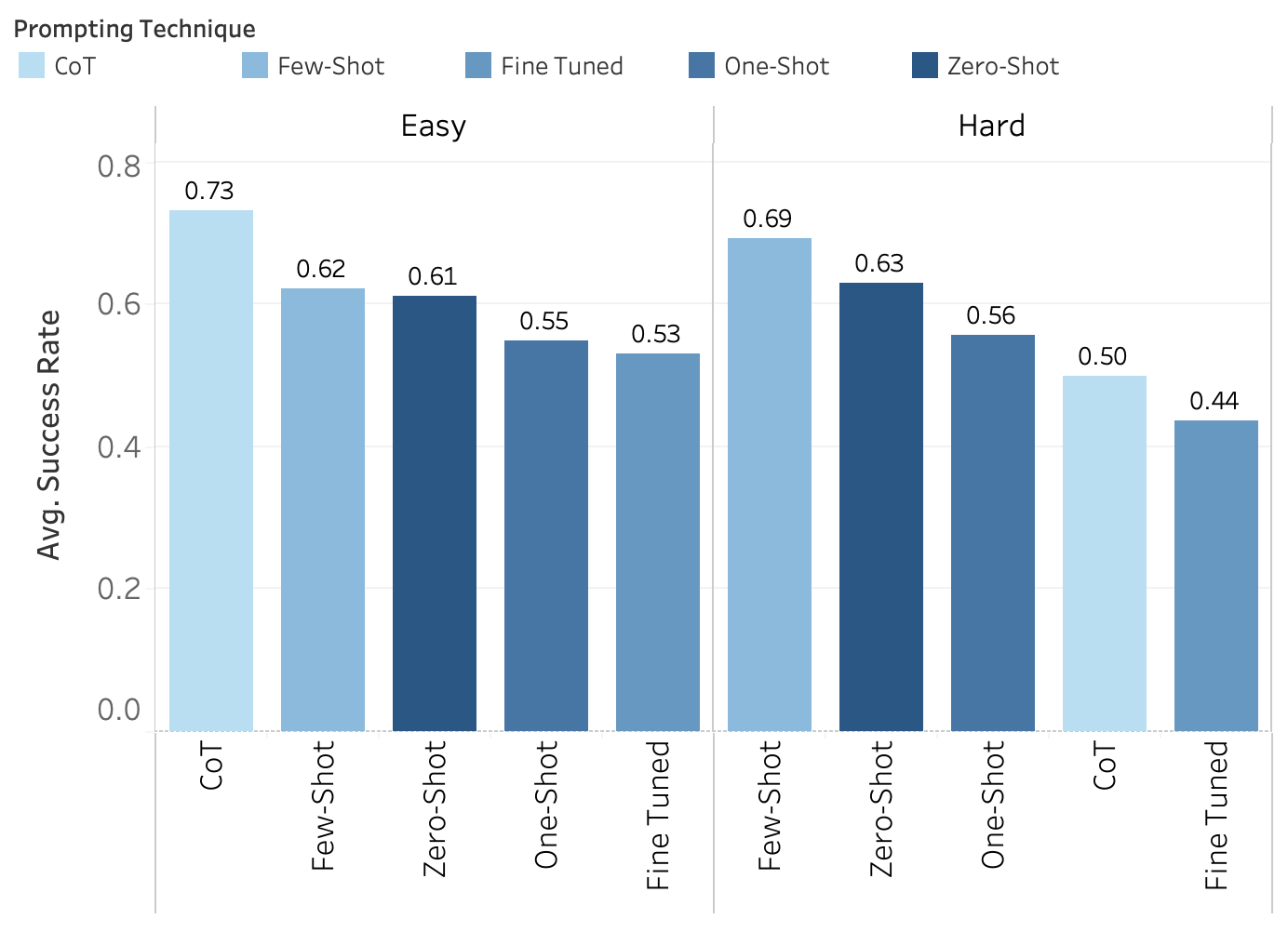}
            \caption{Success rate in generating a specific APT based on the human-perceived difficulty of example.}
            \label{fig:difficulty}
        \end{figure}

        To answer this question, we asked our annotators, for each presented generation, whether they would find applying the paraphrase type to a given sentence \textit{easy} or \textit{hard}. 
        Then, we computed the rate of successfully generated paraphrases for different approaches depending on the estimated difficulty of the task, as shown in \Cref{fig:difficulty}. 
        On the x-axis, the tasks are split according to rated difficulty, and the y-axis gives the rate of successful application.
        
        \looseness=-1We observe that the few-shot approach works better in cases where humans evaluate the task as hard, seeming to profit from a human perspective more difficult task. 
        The CoT-prompted model performs poorly for hard examples, with 23\% points less generation success.
        One reason might be that human annotators evaluate examples as hard, which are difficult to reason about, and this extends to reasoning attempts from LLMs.
        Similarly, the fine-tuned model also performs worse for hard examples.        
        Surprisingly, the performance of the other prompt paradigms seems largely independent of the difficulty. 
        The results suggest that explicit reasoning about paraphrasing tasks that are hard for humans is also hard for LLMs. 
        If more complex insights are required, as judged by humans, they fail more often as they lack the required understanding of the concepts associated with the APTs.
        Still, the LLMs exhibit some reasoning capability for APTs, even if they are led astray by it in difficult scenarios. %
  
\begin{figure}[t]
    \centering
    \includegraphics[width=\columnwidth]{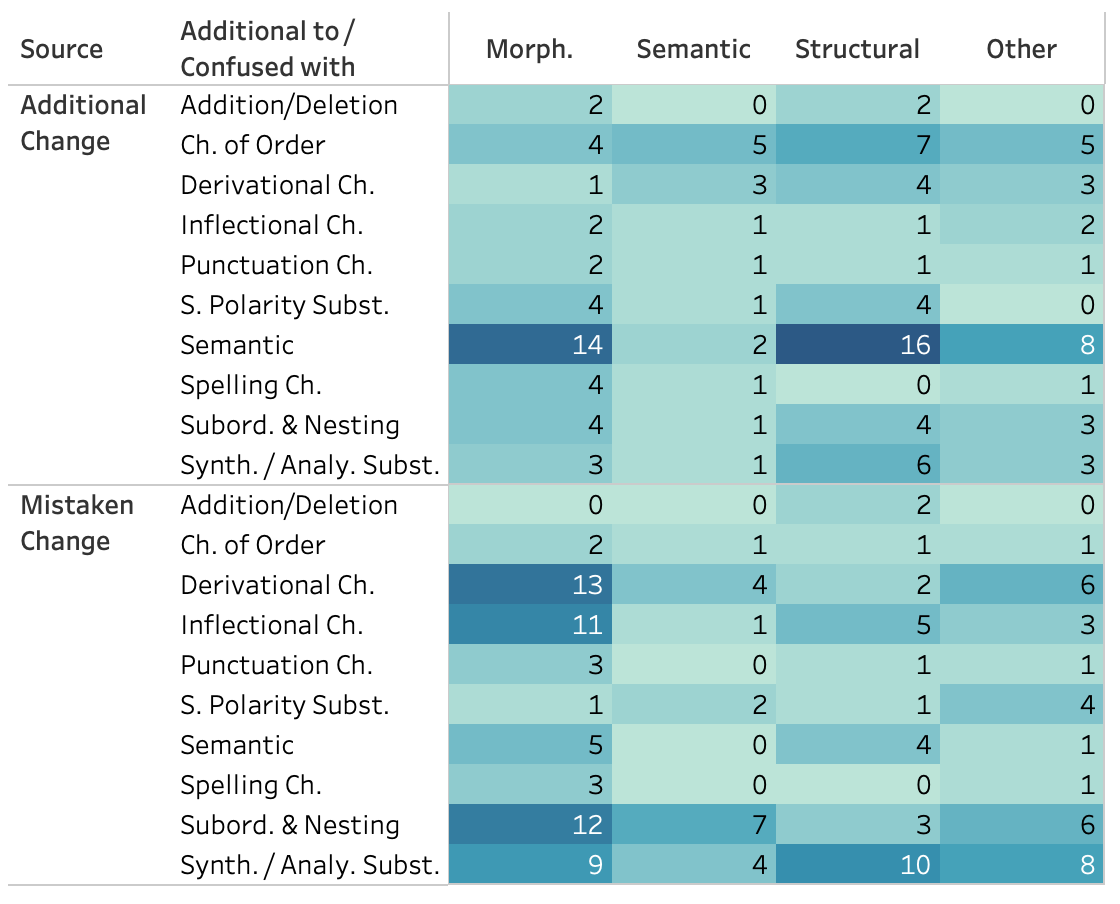}
    \caption{Detailed confusion matrix with absolute values, the maximum is 50 in each case. The column gives the intended APT.}
    \label{fig:detailed_confusion}
\end{figure}

  \questiona{How does the detailed confusion matrix look? Do specific APTs get confused more often than other APTs?}
  \ans To get a more granular view of the confusion, we also looked at the detailed confusion based on single APTs as sources instead of groups.
  The result is shown in \Cref{fig:detailed_confusion}. 
  We plotted the absolute numbers of additional/erroneous changes, as here, all groups are the same size, i.e., for each APT, there are 50 generations.
  We have already noticed that most confusion happens intra-group and not inter-group.
  There are big differences inside the groups, depending on the APT.
  For instance, for morpho-lexical changes, \apt{Same Polarity Substitution} is rarely confused, while \apt{Inflectional and Derivational Changes} are often confused for different morpho-lexical APTs.
  As \apt{Same Polarity Substitution}, i.e., the change of one unit for a synonym, is a very common APT, the model seems to have a very good understanding of it.
  Still, additional changes to \apt{Same Polarity Sub.} are made at a comparable or even larger rate than other group members.
  So, even with a good representation in the trainings data, it is difficult for the model to generate only the requested type.
  The big intra-group differences suggest that it makes sense to characterize the common types in as much detail as possible for any paraphrasing-related task, as model performance differs at the APT group and the APT level.
    
\begin{figure}
    \centering
    \includegraphics[width=\columnwidth]{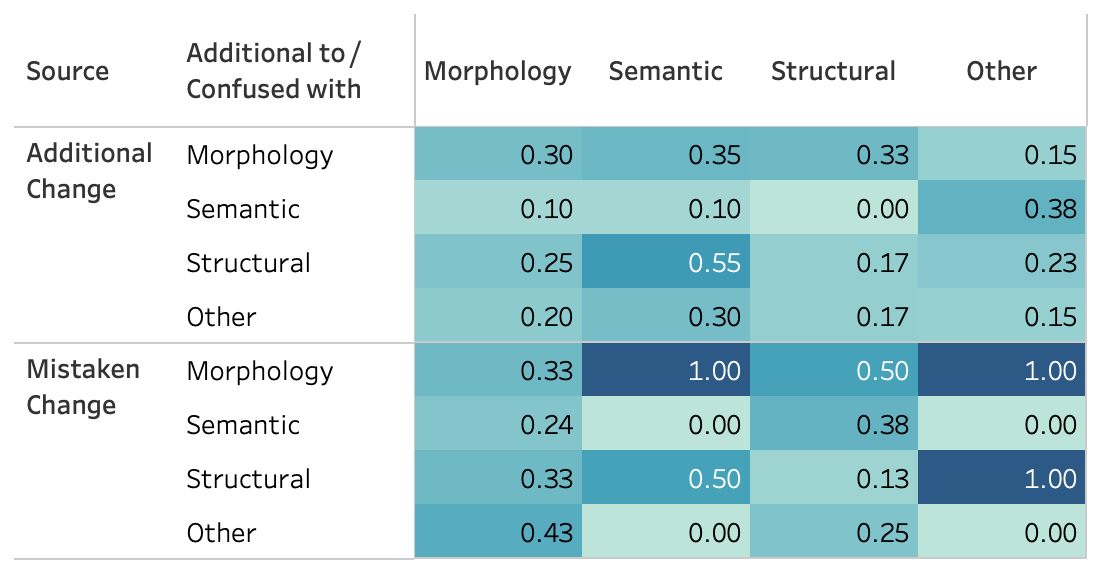}
    \caption{Confusion matrix only for hard examples. The columns give the intended APT.}
    \label{fig:confusion_hard}
\end{figure}

\questiona{Is there a correlation between perceived hardness of a task and confusion?}
\ans We have already looked at how difficulty relates to model performance regarding generation success.
    We also wanted to investigate how it relates to confusion.
    Therefore, we plotted the same confusion matrix for tasks rated as hard in \Cref{fig:confusion_hard}. 
    It follows similar trends to the unrestricted confusion matrix but with remarkably higher confusion rates across the board.
     
    This further supports the assumption that tasks rated hard by humans are also harder for LLMs.
    If a task is rated hard, \gptshort\space is likelier to perform additional, unwanted changes and even more likely to perform the wrong kind of change.

 \questiona{How much do humans and \gptshort\space agree on paraphrase type evaluation?}
    \ans Human annotation at scale is costly and difficult to perform.
    Works like \cite{huangHumanAnnotatorsChatGPT} have raised the possibility of using LLMs for annotation tasks.
    Besides potentially posing as an alternative to human annotation, LLM annotation might also give insight into how well LLMs understand paraphrase types from a detection perspective.
    Therefore, we want to see how well LLM annotation aligns with human annotations.
    To that end, we used the same annotation document provided to the annotators to prompt \gptshort\space and asked it to perform the same tasks the human annotators performed.

    We found little agreement between the human annotation and the answers produced by \gptshort. 
    For evaluating whether the given text pair is a paraphrase, the annotations of humans and \gptshort\space are identical in 54 \% of the cases.
    Similarly, to check whether a paraphrase type is applied correctly, humans and \gptshort\space only agree in 43\% of the cases. 
    In evaluating task difficulty, they only agreed in 50 \% of the cases.
    As these annotation tasks were binary choices, i.e., yes or no and easy or hard, \gptshort\space seems to only agree by chance with the human annotation (even with the same instructions).
   
    While an improvement in alignment is probably possible by different prompting techniques and prompts more tailored to the setting, the results still show that \gptshort\space has trouble understanding the phenomena from a detection perspective.
    The models can also not estimate the task difficulty from a human perspective, i.e., \gptshort\space can not estimate what tasks humans find complex or easy.

    \subsection{Result Details}

    The precise values for the success rates corresponding to \Cref{fig:success} of the different prompt methods are given in \Cref{tab:detail_groups}.
    
    \tabcolsep=0.11cm
        \begin{table*}[t]
            \small
            \centering
            \begin{tabular}{llrrrrrr}
\toprule
Group    & APT                       & CoT           & Few-Shot     & Fine-Tuned   & One-Shot     & Zero-Shot    \\
\midrule
Morph.   & Change of Order           & 0.8           & \textbf{0.9} & 0.7          & \textbf{0.9} & 0.8          \\
         & Derivational Change       & \textbf{0.6}  & 0.4          & 0.3          & 0.5          & 0.5          \\
         & Inflectional Change       & \textbf{0.6}  & \textbf{0.6} & \textbf{0.6} & 0.0          & 0.2          \\
         & Spelling Change           & \textbf{0.7}  & \textbf{0.7} & 0.5          & 0.6          & 0.5          \\
         & Same Polarity Subst. C.   & \textbf{1.0}  & 0.9          & 0.3          & \textbf{1.0} & 0.7          \\
Struct.  & Punctuation Change        & \textbf{0.8}  & 0.6          & 0.7          & 0.4          & 0.7          \\
         & Subordination and Nesting & 0.2           & \textbf{0.5} & 0.2          & \textbf{0.5} & \textbf{0.5} \\
Semantic & Semantic-Based Change     & \textbf{1.0}  & \textbf{1.0} & 0.5          & 0.7          & 0.9          \\
Other    & Addition/Deletion         & \textbf{1.0}  & 0.5          & 0.5          & 0.8          & 0.8          \\
         & Synthetic/Analytic Subst. & 0.2           & 0.2          & \textbf{0.7} & 0.1          & 0.5          \\
Average  &                           & \textbf{0.69} & 0.63         & 0.5          & 0.55         & 0.61  \\   
\bottomrule
\end{tabular}

            \caption{The success rate in generating a specific APT for different prompting techniques of \gptshort. The most successful model(s) are marked in \textbf{bold} for each APT.}
            \label{tab:detail_groups}
    \end{table*}

\section{Prompts}
\label{sec:prompts}

In the following, we present the prompts used to generate the paraphrases. The placeholder \{definition\} stands for the definition of an APT, \{sentence\} stands for the base sentence that should be altered, \{example\} is replaced by an example application of the given APT, and lastly, \{type\} is a stand-in for the name of the APT. The examples are constructed from paraphrase pairs in ETPC by manually editing a pair such that only one change of type $l$ is present.

The following figures show the prompt template for zero-shot (\Cref{fig:ap_prompt_template_zeroshot}), one-shot (\Cref{fig:ap_prompt_template_oneshot}), few-shot (\Cref{fig:ap_prompt_template_fewshot}) and chain-of-thought (\Cref{fig:ap_prompt_template_cot}) prompts. The fine-tuned models are prompted according to \cite{wahle-etal-2023-paraphrase} using the sentence and the types. 
Lastly \Cref{fig:ap_prompt_oneshot_example} shows one concrete one-shot prompt and the corresponding model response.

\begin{figure*}[t]
    \begin{AIbox}{Zero-Shot Prompt Template}
    \parbox[t]{\textwidth}{
        In this task you will be given a definition of an alteration and an input sentence in '''. Your task is to apply the alteration to the given sentence, while preserving the original meaning of the sentence. That means, the new sentence should be a paraphrase of the old sentence. Output the altered sentence and check that the nothing except the alterations you made was changed in the sentence and the alteration given is indeed applied.\\
        
        Alteration: \{definition\}\\
        
        Input: '''\{sentence\}'''
    }
    \end{AIbox}
    \caption{The zero-shot template to paraphrase sentences.}
    \label{fig:ap_prompt_template_zeroshot}
\end{figure*}

\begin{figure*}[t]
    \begin{AIbox}{One-Shot Prompt Template for APT}
    \parbox[t]{\textwidth}{
        In this task you will be given a definition of an alteration, an example of the alteration applied to a sentence and an input sentence in '''. Your task is to apply the alteration to the given sentence, while preserving the original meaning of the sentence. That means, the new sentence should be a paraphrase of the old sentence. Output the altered sentence and check that the nothing except the alterations you made was changed in the sentence and the alteration given is indeed applied.\\
        
        Alteration: \{definition\}\\
        
        Example: \{example\}\\
        
        Input: '''\{sentence\}'''
        
    }
    \end{AIbox}
    \caption{The one-shot template to paraphrase sentences.}
    \label{fig:ap_prompt_template_oneshot}
\end{figure*}

\begin{figure*}[t]
    \begin{AIbox}{Example One-Shot Prompt}
    \parbox[t]{\textwidth}{
        In this task you will be given a definition of an alteration, an example of the alteration applied to a sentence and an input sentence in '''. Your task is to apply the alteration to the given sentence, while preserving the original meaning of the sentence. That means, the new sentence should be a paraphrase of the old sentence. Output the altered sentence and check that the nothing except the alterations you made was changed in the sentence and the alteration given is indeed applied.\\
        
        Alteration: \\Addition/Deletion: Addition/Deletion consists of all additions/deletions of lexical and functional units.\\
        
        Example:\\ Input: '''Amrozi accused his brother, whom he called "the witness", of deliberately distorting his evidence.''' \\
        Output: Amrozi accused his brother, whom he only called "the witness", of deliberately distorting his evidence.\\
        
        Input: '''They had published an advertisement on the Internet, offering the cargo for sale, he added.'''

        \par\noindent\rule{\textwidth}{0.4pt}
        Model Output: "\hl{They had published an advertisement on the Internet on June 10, offering the cargo for sale, he added.}"
    }
    \end{AIbox}
    \caption{Specific example for the one shot with APT were the \hl{highlighted} font indicates the model response.}
    \label{fig:ap_prompt_oneshot_example}
\end{figure*}

\begin{figure*}[t]
    \begin{AIbox}{Few-Shot Prompt Template}
    \parbox[t]{\textwidth}{
        In this task you will be given a definition of an alteration, examples of the alteration applied and an input sentence in '''. Your task is to apply the alteration to the given sentence, while preserving the original meaning of the sentence. That means, the new sentence should be a paraphrase of the old sentence. Output the altered sentence and check that the nothing except the alterations you made was changed in the sentence and the alteration given is indeed applied.\\
        
        Alteration: \{definition\}\\
        
        Examples:\\
        \hspace*{5mm}\{example\}\\
        \hspace*{5mm}\{example\}\\
        \hspace*{5mm}\{example\}\\
        \hspace*{5mm}\{example\}\\
        \hspace*{5mm}\{example\}\\
        
        Input: '''\{sentence\}'''
    }
    \end{AIbox}
    \caption{The few-shot template to paraphrase sentences.}
    \label{fig:ap_prompt_template_fewshot}
\end{figure*}

\begin{figure*}[t]
    \begin{AIbox}{Chain-of-Thought Prompt Template}
    \parbox[t]{\textwidth}{
        In this task you will be given a definition of an alteration and an input sentence in '''. Your task is to apply the alteration to the given sentence, while preserving the original meaning of the sentence. That means, the new sentence should be a paraphrase of the old sentence. Think step by step and describe the reason for what part of the sentence you are changing before you do. Output the altered sentence at the end in the format given below, that is with "Output: " in front.\\

        Alteration: \{definition\}\\
        
        Example: \{example with explanations\}\\
        
        Input: '''\{sentence\}'''
    }
    \end{AIbox}
    \caption{The chain-of-thought template to paraphrase sentences.}
    \label{fig:ap_prompt_template_cot}
\end{figure*}

\section{APT Definitions}
\label{sec:apt_def}
\apt{Addition/Deletion}: Addition/Deletion consists of all additions/deletions of lexical and functional units. \\
\apt{Same Polarity Substitution (contextual)}: Same Polarity Substitution consists of changing one lexical (or functional) unit for another with approximately the same meaning. Among the linguistic mechanisms of this type, we find synonymy, general/specific substitutions, or exact/approximate alternations. \\
\apt{Syntheticanalytic substitution}: Synthetic/analytic substitution consists of changing synthetic structures for analytic structures, and vice versa. This type comprises mechanisms such as compounding/ decomposition, light element, or lexically emptied specifier additions/deletions, or alternations affecting genitives and possessive. \\
\apt{Change of order}: Change of order includes any type of change of order from the word level to the sentence level.\\
\apt{Punctuation changes}: Punctuation and format changes consist of any change in the punctuation or format of a sentence (not of a lexical unit, cf. lexicon-based changes). \\
\apt{Inflectional Changes}: Inflectional changes consist of changing inflectional affixes of words \\
\apt{Spelling changes}: Spelling and format changes comprise changes in the spelling and format of lexical (or functional) units, such as case changes, abbreviations, or digit/letter alternations. \\
\apt{Subordination and nesting changes}: Subordination and nesting changes consist of changes in which one of the members of the pair contains a subordination or nested element, which is not present, or changes its position and/or form within the other member of the pair. \\
\apt{Semantic based Changes}: Semantics-based changes are those that involve a different lexicalization of the same content units. These changes affect more than one lexical unit and a clear-cut division of these units in the mapping between the two members of the paraphrase pair is not possible. \\
\apt{Derivational Changes}: Derivational Changes consist of changes of category with or without using derivational affixes. These changes imply a syntactic change in the sentence in which they occur. \\

\section{Annotation Guidelines}
\label{app:guidelines}
Dear annotators,
In this experiment, we want to explore how humans perceive generated paraphrases. Paraphrases are texts expressing identical meanings that use different words or structures. However,  instead of looking at general paraphrases, we want to examine specific paraphrase types. Paraphrase types, also known as atomic paraphrase types, are specific lexical, syntactic, and semantic changes that can be grouped into a hierarchical topology. 
Each annotator will work on roughly 40 examples. 

Your role will be to look at paraphrases and evaluate if these generations were successful.

Disclaimer: No sensitive information will be collected during annotation.

\subsection{Paraphrases - Background}

Paraphrasing is the act of rephrasing or restating a text or idea using different words while retaining the original meaning. For example, consider the following two sentences:

    1. \color{blue}Amrozi accused his brother \color{black}, \color{violet} whom \color{black} he \color{violet} called \color{black} "the witness", of deliberately distorting his evidence.
    2. \color{violet} Referring to him \color{black} as \color{teal} only \color{black} "the witness", \color{blue} Amrozi accused his brother \color{black} of deliberately distorting his evidence.

You can see that these two sentences, while using a different order and wording and possibly expressing different nuances, mean the same thing. We are specifically interested in the linguistic building blocks that make up paraphrases. These can take different forms (e.g., lexical, semantic). We can illustrate some of them with the example above. The paraphrase types that appear are:

\color{blue}Change of order \color{black}: Change of order includes any type of change of order from the word level to the sentence level. Here, the part “Amrozi accused his brother” is moved in the sentence.

\begin{figure}
    \centering
    \includegraphics[width=1\linewidth]{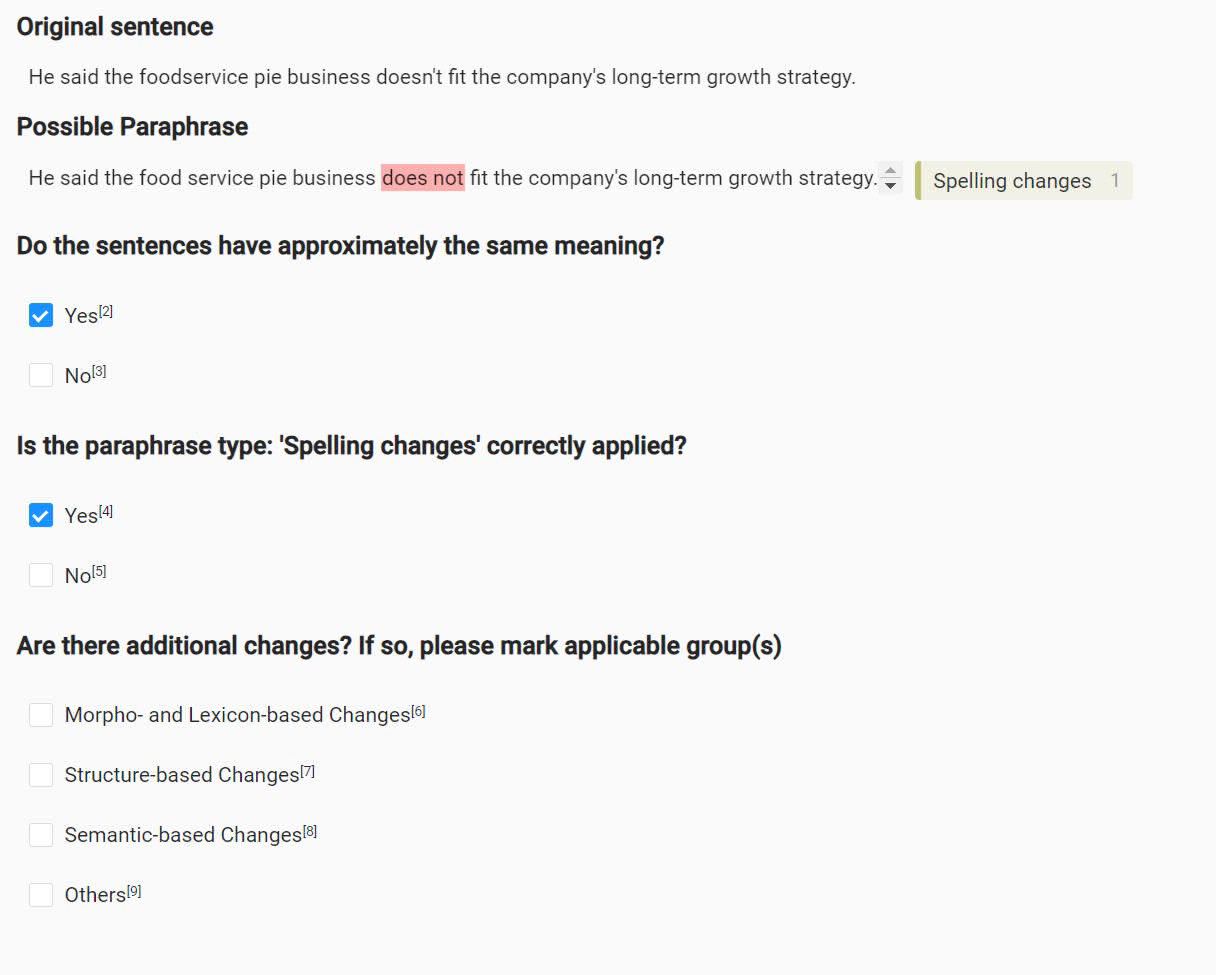}
    \caption{Annotation example interface.}
    \label{fig:anno_example}
\end{figure}

\color{violet} Same Polarity Substitution \color{black}: Same-polarity substitutions change one lexical (or functional) unit for another with approximately the same meaning. In this example, this happens with “whom” and “to him” and “called” and “Referring” respectively.

\color{teal} Addition/Deletion \color{black}: This type consists of all additions/deletions of lexical and functional units. The word “only” is added.

Please find the full list of atomic paraphrase types we will consider at the end of the document.

In \autoref{fig:anno_example}, we show an example of how the annotation process will take place.

\subsection{Tasks}
You will be given a base sentence, an atomic paraphrase type, and an altered sentence. Additionally, you will be given the definition and an example for each atomic paraphrase type you are working on. Please see Figure 1 for an example annotation.
Please read these additional materials carefully to understand what a valid paraphrase of that type would look like.

\begin{enumerate}[label=\Alph*]
    \item Indicate whether the altered sentence is a paraphrase of the base sentence
    \item Indicate whether the altered sentence contains a correct application of the given atomic paraphrase type.
    \begin{enumerate}[label=\arabic*]
        \item If yes:  
        \begin{itemize}
            \item Highlight the given change
        \end{itemize}
        \item If no:
        \begin{itemize}
        \item Please indicate what went wrong with the application of the paraphrase type:
            \begin{enumerate}
                \item The sentences are identical
                \item Nonsense
                \item Different APTs were applied 
                \item Other reason
            \end{enumerate}
         \end{itemize}

        \item If additional or different changes were made than the one initially provided, please identify the groups (see item Atomic Paraphrase Type Groups) of change the additional changes belong to. See the end for a reference of possible categories.
    \end{enumerate}

    \item For each example which you annotated, please indicate whether you find applying the paraphrase to the sentence easy or hard.

    \item Sometimes additional text, besides the paraphrase, might be given (e.g. “Altered sentence” or explanations about the change). Please indicate whether that was the case. Disregard the additional text for the previous tasks.
\end{enumerate}

\textit{The annotation interface will support you as far as possible at the annotation and only show the decisions you need to make depending on your prior annotations, i.e., unnecessary questions will not be displayed.}

In case of any questions whether about the process or any specific example, please contact me at \{author email\}. When you are done with all assigned tasks, please also send me a quick email letting me know.

\subsection{Atomic Paraphrase Types:}
\apt{Addition/Deletion} consists of all additions/deletions of lexical and functional units. The word “only” was added/removed in the example below.
\begin{enumerate}[label=\alph*]
    \item Amrozi accused his brother, whom he called "the witness", of deliberately distorting his evidence.
    \item Amrozi accused his brother, whom he only called "the witness", of deliberately distorting his evidence.
\end{enumerate}

\apt{Same Polarity Substitution} changes one lexical (or functional) unit for another with approximately the same meaning. Among the linguistic mechanisms of this type, we find synonymy, general/specific substitutions, or exact/approximate alterations. 
\begin{enumerate}[label=\alph*]
    \item Apple noted that half the songs were purchased as part of albums.
    \item Apple said that half the songs were purchased as part of albums.
\end{enumerate}
Synthetic/analytic substitution consists of changing synthetic structures for analytic structures, and vice versa. That means that the concept of the predicate is already included in the concept of the subject or the additional predicate is removed. This type comprises mechanisms such as compounding/ decomposition, light element, or lexically emptied specifier additions/deletions, or alterations affecting genitives and possessives.
    a. About 120 potential jurors were being asked to complete a lengthy questionnaire.
    b. The jurors were being asked to complete a lengthy questionnaire.

\apt{Change of order} includes any type of change of order from the word level to the sentence level. See the example in the Background section.
\begin{enumerate}[label=\alph*]
    \item Amrozi accused his brother, whom he called "the witness", of deliberately distorting his evidence
    \item Calling him "the witness", Amrozi accused his brother of deliberately distorting his evidence. 
\end{enumerate}

\apt{Punctuation changes} consist of any change in the punctuation or format of a sentence (not of a lexical unit, like doesn’t to does not).
\begin{enumerate}[label=\alph*]
    \item  PG\&E Corp. shares jumped \$1.63, or 8 percent, to close Friday at \$21.51 on the New York Stock Exchange.
    \item PG\&E Corp. shares jumped \$1.63 or 8 percent to close Friday at \$21.51 on the New York Stock Exchange.
\end{enumerate}

\apt{Inflectional Changes} consist of changing inflectional affixes of words. In the example, a plural/singular alternation (streets/street) can be observed. 
\begin{enumerate}[label=\alph*]
    \item It was with difficulty that the course of streets could be followed 
    \item It was with difficulty that the course of the street could be followed 
\end{enumerate}

\apt{Spelling changes and format changes} comprise changes in the spelling and format of lexical (or functional) units, such as case changes, abbreviations, or digit/letter alterations.
\begin{enumerate}[label=\alph*]
    \item The DVD-CCA then appealed to the state Supreme Court.
    \item The DVD CCA then appealed to the state Supreme Court.
\end{enumerate}

\apt{Subordination and nesting changes} consist of changes in which one of the members of the pair contains a subordination or nested element, which is not present, or changes its position and/or form within the other member of the pair. What is a relative clause in (b) (a spokeswoman for Child) is part of the main clause in Example (a)
\begin{enumerate}[label=\alph*]
    \item  Sheena Young of Child, the national infertility support network, hoped the guidelines would lead to a more "fair and equitable" service for infertility sufferers.
    \item Sheena Young, a spokesman for Child, the national infertility support network, hoped the guidelines would lead to a more "fair
\end{enumerate}

\apt{Semantic based changes} are those that involve a different lexicalization of the same content units. These changes affect more than one lexical unit and a clear-cut division of these units in the mapping between the two members of the paraphrase pair is not possible. In the example, the content units referring to increases are present in both sentences, but there is not a clear-cut mapping between the two. 
\begin{enumerate}[label=\alph*]
    \item The largest gains were seen in prices, new orders, inventories and exports. 
    \item Prices, new orders, inventories and exports increased.
\end{enumerate}

 \apt{Derivational Changes} consist of changes of category with or without using derivational affixes. These changes imply a syntactic change in the sentence in which they occur.
\begin{enumerate}[label=\alph*]
    \item  Tyco later said the loan had not been forgiven, and Swartz repaid it in full, with interest, according to his lawyer, Charles Stillman.
    \item Tyco later said the loan had not been forgiven, and Swartz repaid it fully, with interest, according to his lawyer, Charles Stillman.
\end{enumerate}

\subsection{Atomic Paraphrase Type Groups}
\label{sec:apt_groups}

\textbf{Morpholexical Changes}: These include all changes where a single word or lexical unit is changed. From the paraphrase types you have seen, this includes:
\begin{enumerate}[label=\alph*]
    \item Inflectional Changes 
    \item Derivational Changes
    \item Spelling changes and format changes
    \item Same Polarity Substitution
    \item Synthetic/analytic substitution
\end{enumerate}

\textbf{Structure-based Changes}: These include all changes that arise from a different structural organization of a sentence. For examples from the paraphrase types you have seen, this includes:
\begin{enumerate}[label=\alph*]
    \item  Subordination and nesting changes
    \item  Punctuation changes
\end{enumerate}

\textbf{Semantic-based Changes}: These include all changes that arise from distributing semantic meaning across different lexical units. 
\begin{enumerate}[label=\alph*]
   \item  Semantic Based Changes
\end{enumerate}
\textbf{Others}: Any other changes. For examples from the paraphrase types you have seen, this includes:
\begin{enumerate}[label=\alph*]
    \item  Change of Order
    \item  Addition/Deletion
\end{enumerate}

\section{Tool Use Acknowledgments}
In the conduct of this research project, we used specific artificial intelligence tools and algorithms: ChatGPT, Gemini and DeepL Write to assist with phrasing and editing. While these tools have augmented our capabilities and contributed to our findings, it's pertinent to note that they have inherent limitations. We have made every effort to use AI in a transparent and responsible manner. Any conclusions drawn are a result of combined human and machine insights. This is an automatic report generated with © AI Usage Cards https://ai-cards.org
\newpage
\hypertarget{annotation}{}
\citationtitle

\begin{bibtexannotation}
@inproceedings{meier-etal-2025-towards,
 author={Meier, Dominik, Wahle, Jan Philip, Lima Ruas, Terry, Gipp, Bela},
 title={Towards Human Understanding of Paraphrase Types in Large Language Models},
 address={Abu Dhabi, UAE},
 booktitle={Proceedings of the 31st International Conference on Computational Linguistics},
 editor={Rambow, Owen, Wanner, Leo, Apidianaki, Marianna, Al-Khalifa, Hend, Eugenio, Barbara Di, Schockaert, Steven},
 pages={6298--6316},
 publisher={Association for Computational Linguistics},
 year={2025},
 month={01}
}\end{bibtexannotation}

\end{document}